\documentclass[10pt]{JournalTitle}
\usepackage{graphicx}
\usepackage[singlelinecheck=off,font={footnotesize,bf},labelsep=space]{caption}
\usepackage{float}
\usepackage{amsmath}
\usepackage{amssymb}
\usepackage{CJKutf8}
\usepackage{bm}
\usepackage{multirow}
\usepackage[misc]{ifsym}

\title{Converse Attention Knowledge Transfer for Low-resource Named Entity Recognition}
\titleshort{Converse Attention Knowledge Transfer for Low-Resource Named Entity Recognition}
\author{
	Shengfei Lyu\footnotemark[1], Linghao Sun\footnotemark[1], Huixiong Yi\footnotemark[1],
	Yong Liu\footnotemark[2],
	Huanhuan Chen\footnotemark[1],
	Chunyan Miao\footnotemark[2]}

\footnotetext[1]{University of Science and Technology of China, Hefei 230027, China.}
\footnotetext[2]{Nanyang Technological University,  Nanyang Ave 639798, Singapore. \par
\hskip-7pt{Correspondence Author: Shengfei Lyu (saintfe@mail.ustc.edu.cn).}}

\doi{{\quad}}
\volume{X}
\page{1--10}

\abstract{In recent years, great success has been achieved in many tasks of natural language processing (NLP), e.g., named entity recognition (NER),
	especially in the high-resource language, i.e., English, thanks in part to the considerable amount of labeled resources. 
	However, most low-resource languages do not have such an abundance of labeled data as high-resource English, 
	leading to poor performance of NER in these low-resource languages.
	Inspired by knowledge transfer, we propose  Converse Attention Network, or CAN in short, 
	to improve the performance of  NER in low-resource languages by leveraging the knowledge learned in pretrained high-resource English models. 
	CAN first translates low-resource languages into high-resource English using an attention based translation module.
	In the process of translation, CAN obtain the attention matrices that align the two languages. 
	Furthermore, CAN use the attention matrices to align the high-resource semantic features from a pretrained high-resource English model with the low-resource semantic features.
	As a result, CAN obtains aligned high-resource semantic features to enrich the  representations of low-resource languages. 
	Experiments on four low-resource NER datasets show that CAN achieves consistent and significant performance improvements, 
	which indicates the effectiveness of CAN.}


\begin{document}
\maketitle
	
	\begin{multicols}{2}
		\begin{figure*}
			\centering
			\includegraphics[width=1\textwidth]{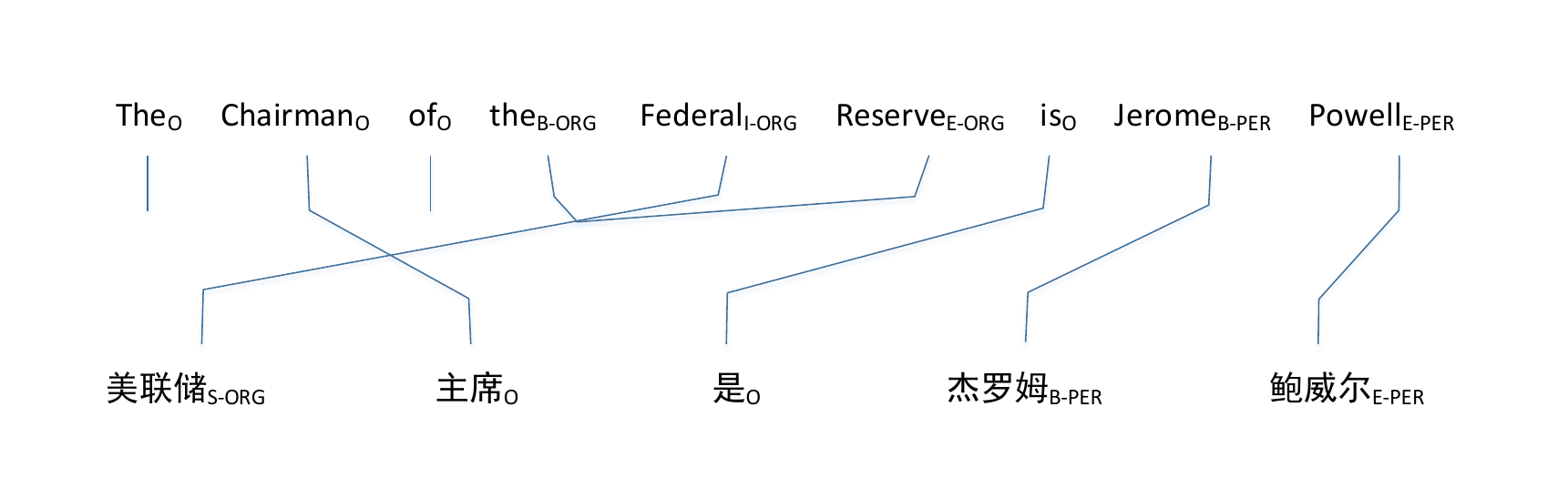}
			\caption{Examples of word-aligned bilingual (English and Chinese) parallel sentences for NER in BIOES format.} 
			\label{example1}
		\end{figure*}
		{N}amed entity recognition (NER)  is a fundamental task in natural language processing (NLP), 
		which benefits many applications, such as information extraction~\cite{lyu-chen-2021-relation,lyu-etal-2022-auxiliary,lyu-etal-2022-rdr,hu2022contextual}
		and question answering~\cite{zhao-2020-kdd-condition,zhao-2021-tnnls}, in financial and medical fields.
		NER is usually regarded as a sequence tagging task 
		that classifies  continuous tokens into specified categories, 
		such as \textit{persons}, \textit{organizations}, and \textit{locations}. 
		The state-of-the-art NER methods usually employ long short-term memory recurrent neural networks (LSTM) 
		and a subsequent conditional random field (CRF) to tag tokens of a sequence~\cite{huang2015bidirectional}. 
		Due to the adoption of deep neural networks, these methods require large-scale labeled data to be trained.
		
		Generally, English is regarded as the  high-resource language, while other languages, even Chinese, are treated  as low-resource languages~\cite{feng2018improving}.
		As English is widely used and studied in the world, 
		abundant English labeled NER data are available for training large models~\cite{sang2003introduction, weischedel2011ontonotes}.
		By the way, NER in the high-resource  language (i.e., English) is also called high-resource NER.
		English NER models~\cite{akbik2019pooled} with good performance are pretrained thanks in part to these sufficient labeled  resources. 
		By contrast, the languages other than English are still not fully studied due to the lack of labeled data.
		The insufficiency of labeled data is partly due to the fact that manually labeling data is expensive and time-consuming and 
		many institutions and/or researchers with limited resources are hard to afford large amounts of high-quality labeled data. 
		As a result, performance of neural NER models in these low-resource languages is compromised since the training data are insufficient~\cite{zhang2016name}.

		Recently, many cross-lingual learning methods were applied to address the  NER task on low-resource languages (aka low-resource NER). 
		Previous work was mostly based on heuristic methods and transferred words between two languages~\cite{wang2014cross}. 
		For example, Che \textit{et al.}~\cite{che2013named} predicted entity labels on labeled parallel datasets of two languages.
		Words in the two languages are aligned to form word pairs. 
		Then, two NER models trained in the two languages were constrained to be consistent in the form of joint prediction of word pairs.
		Although these methods achieved good performance, they required labeled parallel corpora, 
		which are more expensive than monolingual labeled corpora~\cite{che2013named}. 
		Without labeling parallel corpora, some methods utilized translation models to establish an alignment between the low-resource and high-resource languages. 
		For example, Feng \textit{et al.}~\cite{feng2018improving} enriched the embedding of a source word with the embedding of its translated word in another language.
		The core idea is to exploit the alignment between words in two languages.

		However, it is often difficult to align words precisely in two languages. 
		Generally, it is often that two sentences in two languages share the same meaning but they have different number of words.
		For example, as shown in Figure~\ref{example1}, 
		the sentence ``The chairman of the Federal Reserve is Jerome Powell" in English comprises 9 words, 
		while its corresponding version in Chinese is 
		``\begin{CJK}{UTF8}{gbsn}美联储主席是杰罗姆鲍威尔\end{CJK}", 
		which has 12 words.
		It is impossible to align the words one by one in the two sentences.
		In addition, we observe that the word orders in the two languages are also different in this example.
		Therefore, it is hard to find a general rule to align words in the two languages. 
		
		Besides, the tags in the two languages are difficult to align. 
		In Figure~\ref{example1}, \textit{the Federal Reserve} is composed of three English words, which are tagged as B-ORG, I-ORG, and E-ORG, respectively, 
		while in Chinese, the Federal Reserve is represented by one Chinese word 
		``\begin{CJK}{UTF8}{gbsn}美联储\end{CJK}", 
		which is tagged as S-ORG.  
		Even if words in the two languages can be aligned, the tags in one language cannot be directly projected to the other language.
		
		In a word, misalignment between words and misalignment between tags in two languages impede 
		knowledge transfer based on direct word-to-word translation from high-resource NER to low-resource NER.
		
		To address the above issue, we propose Converse Attention Network (CAN) to enhance knowledge transfer from high-resource NER to low-resource NER. 
		CAN first translates a low-resource language to English via the attention-based translation model~\cite{gehring2017convolutional} 
		whose attention matrices generated in its encoder-decoder attention layers align source and target languages.
		CAN uses the attention matrices generated in the source and target sentences to 
		align  high-resource features and  low-resource features\footnotemark[3], 
		implementing knowledge transfer from high-resource NER to low-resource NER.
		\footnotetext[3]{To be consistent with the terminology of translation models, 
		low-resource sentences correspond to  source sentences \par 
		while high-resource  sentences correspond to target sentences.}
		CAN is named because of the converse use of attention matrices from target sentences to source sentences.
		Besides, instead of transferring the embedding of the translated word in high-resource NER to 
		the embedding of a source word in low-resource NER, where word misalignment occurs, 
		CAN transfers high-level features
		 (i.e., the outputs of BiLSTM~\cite{graves2005framewise} ) of the high-resource sentence to the low-resource features of the corresponding low-resource sentence via the aforementioned attention matrices.
		Here, we treat word embeddings as low-level features and regard as high-level features the outputs 
		that a neural network takes the word embeddings as input and produces.
		Compared with shared representation methods~\cite{bharadwaj2016phonologically}, 
		CAN has the advantage of leveraging semantic and task-specific features obtained by a pretrained NER model 
		and does not need to find universal features in different languages. 
		In addition, the  high-resource semantic features can be naturally aligned by the encoder-decoder attention matrices of the translation model~\cite{gehring2017convolutional}  without additional processing. 
		Note that CAN does not require any hand-craft features and labeled parallel corpora, which are expensive for low-resource NER.
		
		The contributions of this paper are summarized as follows:
		\begin{itemize} 
			\item We propose a novel method, CAN, to transfer high-level semantic features 
			from the high-resource NER to low-resource NER via attention matrices obtained 
			from an attention-based translation model.
			CAN exploits the attention matrices to address word misalignment and tag misalignment between high-resource NER and low-resource NER. 
			\item Extensive experiments on four datasets from two language families  empirically show that 
			CAN improves performance of low-resource NER, which indicates the effectiveness of CAN.
			
		\end{itemize}

		The rest of the paper is organized as follows. Section~\ref{sec:relted-work}
		introduces related work. The proposed method is presented in Section~\ref{sec:ban}. 
		In Section~\ref{sec:experiments}, 
		the experimental results of the NER task in three different languages (i.e., German, Spanish, and Chinese) are presented.
		Section~\ref{sec:analysis} further analyses the effectiveness of the proposed method CAN and 
		the impact of different attention layers on CAN.
		Section~\ref{sec:conclusion} concludes the paper and discusses the future work.
		
		\section{Related Work}
		\label{sec:relted-work}
		\subsection{Named entity recognition}
		
		NER is the task of recognizing mentions from text and classifying them into predefined semantic types, 
		such as \textit{person}, \textit{organization}, \textit{location}. 
		Existing methods usually treat NER as a sequence labeling problem. 
		Various sequence labeling models, such as hidden Markov models (HMM)~\cite{florian2003named}, 
		conditional random fields (CRF)~\cite{lafferty2001conditional}, 
		have achieved decent performance before deep learning is widely used.
		With the widespread use of deep learning, many neural based methods have been proposed, 
		such as Long Short-Term Memory networks (LSTM)~\cite{hammerton2003named}, 
		LSTM-CRF~\cite{lample2016neural,liu2018empower,peters2017} where CRF is utilized on top of LSTM. 
		Convolutional neural networks (CNN) have also been used on the NER task. 
		Collobert {et al.}~\cite{collobert2011natural} exploited a CNN-CRF structure for NER.
		Ma {et al.}~\cite{ma-hovy-2016-end} proposed a neural network architecture, called LSTM-CNN-CRF, combing LSTM and CNN. 
		Strubell {et al.}~\cite{strubell2017fast} applied Iterated Dilated Convolutional Neural Networks (ID-CNNs) to NER. 
		Furthermore, sentences are treated as sequences of characters by neural networks.
		This way has two distinct advantages.
		Firstly, it is helpful for solving the Out-Of-Vocabulary (OOV) problem. 
		Secondly, it can capture additional morphological and orthographic information. 
		Therefore, Dos {et al.}~\cite{dos2015boosting} applied a character-level CNN to boost a CNN-CRF model. 
		Recently, Zhang and Yang~\cite{zhang2018chinese} proposed a lattice-structured LSTM model, 
		which leveraged word and character information at the same time. 
		They showed that lattice-structured LSTM could make a great improvement on NER. 
		In this paper, the architecture of the proposed method is based on the  widely used LSTM-CRF, 
		and cross-lingual knowledge is utilized to improve the performance of low-resource NER.
		
		\subsection{Cross-lingual learning}
		Cross-lingual learning methods are proposed to address low-resource NER. 
		At present, there are two main types of  methods:
		one is annotation projection based on parallel corpora and 
		the other is the shared representation based on transfer learning. 
		
		\subsubsection{Annotation projection}
		Annotation projection relies on parallel corpora and the alignment of words. 
		The tags of tokens in the high-resource language sentence are projected 
		to their aligned tokens in low-resource languages. 
		Under the cross-lingual setting, many methods are proposed for various NLP tasks, such as POS tagging, parsing, and NER. 
		In the POS tagging task, many representative methods made great improvements~\cite{das2011unsupervised,tackstrom2013token}. 
		In the parsing task, Hwa {et al.}~\cite{hwa2005bootstrapping} proposed a direct projection algorithm 
		for syntactic dependency annotation, and  
		Tiedemann~\cite{tiedemann2014rediscovering} utilized cross-lingually harmonized annotation schemes. 
		On the NER task, NER tags are projected within language pairs by annotation projection~\cite{yarowsky2001inducing,zitouni2008mention,ehrmann2011building}. 
		However, annotation projection is heavily dependent on  the alignment of source and target languages,
		whose quality is dependent on the size of the parallel data.  
		The assumption implied in this kind of methods is  correct alignment between source and target languages. 
		Unfortunately, word misalignment and tag misalignment in two languages are inevitable, 
		since the lengths of the two sentences that share the same meaning in two languages are difficult to guarantee consistency.
		This issue impedes the application of annotation projection.

		\subsubsection{Shared representation}
		
		Shared representation relies on universal features, which can be transferred from the high-resource language to a low-resource language. 
		Once a model is trained in the high-resource language using the delexicalized features
		that do not depend on the forms of words, 
		it can be directly applied to a low-resource language. 
		Tackstrom {et al.}~\cite{tackstrom2012cross} enhanced the model by building cross-lingual word clusters. 
		The word clusters were induced by large parallel corpora and used to generate universal features. 
		Bharadwaj {et al.}~\cite{bharadwaj2016phonologically} bridged the high-resource language and low-resource languages through phonemic transcription. 
		Ni {et al.}~\cite{ni2017weakly} proposed to project distributed representations of words  
		into a common space as language independent features. 
		Chaudhary {et al.}~\cite{chaudhary2018adapting} adapted continuous word representations 
		using linguistically motivated sub-word units: phonemes, morphemes, and graphemes. 
		Other methods  built common feature representations 
		include building a bilingual dictionary~\cite{zirikly2015cross,fang2017model}, 
		utilizing Wikipedia data~\cite{kim2012multilingual,tsai2016cross}. 
		Different from obtaining features of the high-resource language, 
		multitask learning is jointly trained across different languages by sharing parameters. 
		For example, Lin \textit{et al.}~\cite{lin2018multi} used shared character and word embeddings 
		which were trained in a multi-task setting to improve the performance of each dataset. 
		The main advantage of shared representation is that it requires minimal dependency on parallel resources. 
		However, this method is currently strongly limited by the fact that it requires a generic feature representation across languages. 
		
		\section{Converse Attention Network}
		\label{sec:ban}
		\begin{figure*}
			\centering
			\includegraphics[width=1\textwidth]{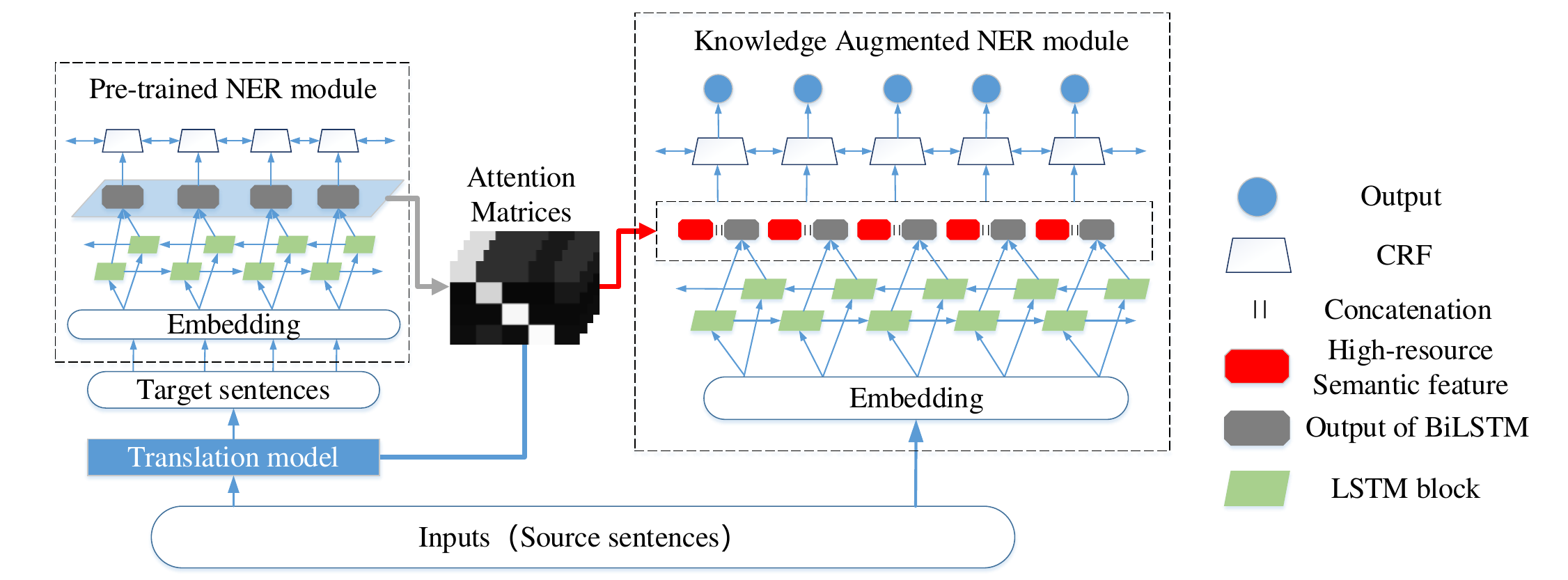}
			\caption{The architecture of CAN. The low-resource sentences are translated into English and CAN records the attention matrices. 
				Then the sentences in the high-resource language are fed into a pretrained model. 
				After acquiring the outputs of BiLSTM in the model, 
				CAN uses the converse attention knowledge transfer mechanism to obtain the aligned high-resource features 
				which are combined with the low-resource features to enrich representations of low-resource words  in the low-resource NER model.} 
			\label{fig:ban}
		\end{figure*}
		
		In this section, the proposed method CAN is introduced in four parts, i.e., attention-based translation module, pretrained NER module, converse attention knowledge transfer, and knowledge augmented NER module. 
		Figure~\ref{fig:ban} illustrates the architecture of CAN.
		
		\subsection{Attention-based translation module}
		
		Following~\cite{gehring2017convolutional}, 
		we use the convolutional sequence to sequence model in the neural machine translation (NMT) module. 
		It divides the translation process into two steps. 
		Firstly, in the encoder step, 
		given an input sentence $\bm{s}=(s_1,\cdots,s_m)$ of length $m$, 
		$\bm{e}^s(s_i)$ maps each word $s_{i}$ to a word embedding $\bm{w}_{i}^{s}$:
		\begin{equation}
			\bm{w}_i^s=\bm{e}^s(s_i),
		\end{equation}
		where $\bm{e}^s$ denotes the word embedding lookup table. 
		After that, the absolute position information of input elements, denoted as positive embeddings $[\bm{p}_1,\cdots,\bm{p}_m]$, is combined with the embedding. 
		Specifically, word embeddings and position embeddings are element-wise added to get sentence representations $[\bm{w}_1^s+\bm{p}_1,\cdots,\bm{w}_m^s+\bm{p}_m]$ as input. 
		Similarly, the embeddings $[\bm{g}_1,\cdots,\bm{g}_n]$ of target words in the decoder network are generated in the same way.
		A convolutional neural network (CNN) is used to extract features of a sentence  from left to right.
		In the decoder step, attention mechanism is used in each CNN layer. 
		In order to acquire the attention value, 
		the $i$th decoder state $\bm{h}^l_i$ in the $l$th level is combined with the embedding of its previous decoder output value $\bm{g}_i$:
		\begin{equation}
			\bm{d}^l_i=\bm{W}^l_d\bm{h}^l_i + \bm{b}^l_d +\bm{g}_i,
		\end{equation}
		where $\bm{W}^l_d$ and $\bm{b}^l_d$ are learned parameters.
		
		For the $l$th encoder-decoder attention layer, the weight $a^l_{ij}$ is computed as a dot-product between the decoder state summary $\bm{d}^l_j$ 
		and the $i$th output (denoted as $\bm{z}_i$) of the encoder block:
		\begin{equation}
			\label{eq:attention}
			a^l_{ij}=\frac{\exp(\bm{d}^l_j \cdot \bm{z}_i)}{\sum_{t = 1}^{m}\exp(\bm{d}^l_j \cdot \bm{z}_t)}.
		\end{equation}
		Following the normal decoder implementation, it gets the target sentence ${t}=(t_1,\cdots,t_n)$ by beam search strategies. 
		
		\subsection{Pre-trained NER module}
		\label{pretrained}
		We use the model proposed in~\cite{akbik2019pooled}, which is one of the state-of-the-art English NER methods. 
		This model utilizes a BiLSTM network as a character-level language model (CharLM) to take contextual information. 
		The hidden states of the character language model are used to create contextualized word embeddings to represent the input words. 
		
		In the forward direction of the character-level language model, 
		the last character of a word is regarded as the word vector, 
		which contains the contextual information from the beginning of the sentence. 
		The backward direction model functions in the same way but in the reversed direction. 
		Formally, we define the forward and backward character embeddings of each word $s_i$ as 
		$\overrightarrow{\bm{h}}_1,\cdots,\overrightarrow{\bm{h}}_l$ and $\overleftarrow{\bm{h}}_1,\cdots,\overleftarrow{\bm{h}}_l$, 
		where $l$ indicates the length of the word.  
		Then, the contextual embedding of the word $s_i$ is represented as follows:
		\begin{equation}
			\bm{e}_i^{c}=[\overrightarrow{\bm{h}}_{l}; \overleftarrow{\bm{h}}_{1} ],
		\end{equation}
		where $\overrightarrow{\bm{h}}_{l}$ and $\overleftarrow{\bm{h}}_{1}$ denote the hidden state of the last character of the word in the forward LSTM and the hidden state of the first character of the word in the backward LSTM, respectively. 
		The concatenation operation is denoted as [;]. 
		
		The final embedding (denoted as $\bm{e}_i$) of the word $s_i$ is formed by 
		concatenating the character-level language model embedding $\bm{e}_i^{c}$ and its GloVe embedding $\bm{e}_i^{g}$~\cite{pennington2014glove}. 
		Namely, $\bm{e}_i = [\bm{e}_i^{c};\bm{e}_i^{g}]$.
		A standard BiLSTM-CRF model takes the embedding $\bm{E}=[\bm{e}_1,\cdots,\bm{e}_n]$ to address the NER task. 
		The English NER model is trained on the CoNLL-2003 English dataset~\cite{sang2003introduction} 
		and the parameters are fixed to predict translated sentences.

		\subsection{Converse attention knowledge transfer}
		
		
		Given an input sentence $\bm{s}=(s_1,\cdots,s_m)$ in a low-resource language, 
		the translation module translates $\bm{s}$ into the high-resource language (i.e., English) and the output sequence is denoted as $\bm{t}=(t_1,\cdots,t_n)$. 
		At the moment, the weights of the encoder-decoder attention layers serve as the bridge of knowledge transfer. 
		

		In the high-resource language,  the BiLSTM output state for the word $t_j$ in  the pretrained English NER model  is:
		\begin{equation}
			\bm{r}_j^t=[ 
			\overrightarrow{\bm{r}^{t}_{j}};
			\overleftarrow{\bm{r}^{t}_{j} }] \in \mathbb{R}^{2d},
		\end{equation}
		where $\overrightarrow{\bm{r}^{t}_{j}}$ and $\overleftarrow{\bm{r}^{t}_{j}}$ denote the outputs of $t_j$ in the forward and backward LSTMs, 
		respectively, and $d$ denotes the dimension of the hidden state of the forward/backward LSTM.  
		We claim that $\bm{r}_j^t$ contains the semantic and task-specific features of the translated sentence. 
		However, it cannot be directly used in the low-resource language, 
		because of the word misalignment between the high-resource language and the low-resource language. 
		
		The encoder-decoder attention matrices imply the alignment of words between the source sentence and the target sentence.
		The weights in an attention layer can serve as a transformation matrix as the bridge of knowledge transfer 
		from the high-resource language to the low-resource language.
		Note that the attention matrix in each attention layer has the same number of rows as the length of the source sentence 
		and the same number of columns as the length of the target sentence. 
		The $i$th row of the attention  matrix, $\bm{a}_{i}^{l}=[a_{i1}^{l},\cdots,a_{in}^{l}]$ in the $l$th attention layer 
		indicates the correlation between the source word $s_i$ with each word in the target sentence in this attention layer.
		Thereafter, the aligned high-resource semantic feature (denoted as $\bm{t}_i^a$) of the $i$th source word $s_i$ is obtained  
		by the weighted sum of the outputs of BiLSTM in the pretrained English model:
		\begin{equation}
			\label{eq:high-feature}
			{\bm{t}^l_i}^{\mathsf{T}} = \sum_{j=0}^{m} a_{ij}^{l}\bm{r}_j^t = \bm{a}_i^{l}\bm{R}^t,
		\end{equation}
		where $\bm{R}^t=[\bm{r}_1^{t},\cdots,\bm{r}_n^{t}]^{\mathsf{T}} \in \mathbb{R}^{n \times 2d}$ 
		represents the whole outputs of BiLSTM in the pretrained English model, and $\bm{a}_i \in \mathbb{R}^{1 \times n}$, 
		$\bm{t}^l_i \in \mathbb{R}^{ 2d}$. 
		Note that $\bm{t}_i^l$ has the same dimension as $\bm{r}_i^t$. 
		
		As for the whole source sentence, the aligned high-resource semantic features (denoted as $\bm{T}^l$) is obtained by the production of
		the attention matrix of the $l$th attention layer and the outputs of BiLSTM in the pretrained English model:
		\begin{equation} \label{eq:high-feature-matrix}
			\bm{T}^l = \bm{A}^{l} \bm{R}^t_{},
		\end{equation}
		where $\bm{T}^l = [{\bm{t}^l_1},\cdots,{\bm{t}^l_m}]^\mathsf{T} \in  \mathbb{R}^{m \times 2d}$ 
		and $\bm{A} = \big[{\bm{a}^l_1}^{\mathsf{T}},\cdots,{\bm{a}^l_m}^{\mathsf{T}}\big]^\mathsf{T} \in  \mathbb{R}^{m \times n}$.

		\subsection{Knowledge augmented BiLSTM-CRF module}
		The  knowledge augmented NER module for low-resource NER is based on the BiLSTM-CRF model. 
		We adopt the same structure as the pretrained English NER model introduced in Subsection~\ref{pretrained}. 
		We enrich the original features obtained by BiLSTM from the low-resource language with  
		the high-resource semantic features (see~(\ref{eq:high-feature-matrix})) to 
		form final representations, which are fed into the last CRF layer. 
		In the following, we depict the proposed method, CAN, in detail.
		
		Firstly, each word in a low-resource language is mapped into a word embedding:
		\begin{equation}
			\bm{w}_i=\bm{e}^s(s_i).
		\end{equation}
		Thus, a sentence $\bm{s}=(s_1,\cdots,s_m)$ in a low-resource language is mapped into $\mathcal{W} = [\bm{w}_1,\cdots,\bm{w}_m]$.
		
		Secondly, BiLSTM takes the sentence $s$ as input to obtain contextual representations of words.
		In each  direction, the representation of each input word is modeled with a single hidden state. 
		Given an initial value, every time step, LSTM consumes an input word and obtains its hidden state recurrently. 
		Take the forward LSTM as an example. 
		the recurrent state denoted as $\overrightarrow{\bm{r}}_k$ for the $k$th word $s_k$  is obtained as follows:
		\begin{equation}
			\begin{aligned}
				&{\bm{i}}_k = \sigma(\bm{W}_i \bm{w}_k + \bm{U}_i \overrightarrow{\bm{r}}_{t-1} + \bm{b}_i) \\
				&{\bm{f}}_k = \sigma(\bm{W}_f \bm{w}_k + \bm{U}_f \overrightarrow{\bm{r}}_{k-1} + \bm{b}_f) \\
				&\bm{o}_k = \sigma(\bm{W}_o \bm{w}_k + \bm{U}_o \overrightarrow{\bm{r}}_{k-1} + \bm{b}_o)\\
				&\bm{u}_k = \tanh(\bm{W}_u \bm{w}_k + \bm{U}_u \overrightarrow{\bm{r}}_{k-1} + \bm{b}_u)\\
				&\bm{c}_k = \bm{c}_{k-1} \odot \bm{f}_k + \bm{u}_k \odot \bm{i}_k\\
				&\overrightarrow{\bm{r}}_k= \bm{o}_k \odot \tanh(\bm{c}_k),
			\end{aligned}
			\label{lstm}
		\end{equation}
		where $\bm{i}_k$, $\bm{o}_k$, $\bm{f}_k$, and $\bm{u}_k$ denote the values of an input gate, an output gate, a forget gate, and an actual input at time step $k$, respectively. These gates control the information flow for a recurrent cell $\bm{c}_k$ and the state vector $\overrightarrow{\bm{r}}_k$. $\bm{W}_x$, $\bm{U}_x$ and $\bm{b}_x$ ($x \in \{i, o, f, u\}$) denotes model parameters. $\sigma$ denotes the sigmoid function.
		The backward LSTM follows the same process as described in~(\ref{lstm}) yet in an opposite direction. 
		The output of the backward LSTM for the word $s_i$ is denoted as $\overleftarrow{\bm{r}}_i$.
		
		Thirdly, the original low-resource representation (denoted as $\bm{r}_k^s$) for the word $s_k$ is obtained by concatenating  $\overrightarrow{\bm{r}}_k$ and $\overleftarrow{\bm{r}}_k$:
		\begin{equation}
			\bm{r}_k^s=[\overrightarrow{\bm{r}}_k;\overleftarrow{\bm{r}}_k] \in \mathbb{R}^{2d}.
		\end{equation}
		
		Finally,  the  low-resource source representation ($\bm{r}_k^s$) 
		and the high-resource semantic representation ($\bm{t}_k^l$ in~(\ref{eq:high-feature})) 
		are concatenated to form the final representation (denoted as $\bm{r}_k^{l}$) for the word $s_k$:
		\begin{equation}
			\bm{r}_k^{l}=[\bm{r}_k^s;\bm{t}_k^l ] \in \mathbb{R}^{4d},
		\end{equation}
		where $l$ denotes the $l$th attention layer whose attention matrix is used to obtain $\bm{t}_k^l$.
		
		Then, a standard CRF layer is utilized on top of the final representation 
		$\bm{R}^{l} = [\bm{r}_1^{l}, \cdots ,\bm{r}_m^{l}]^{\mathsf{T}} \in \mathbb{R}^{m \times 4d}$.
		The state score is obtained  by a linear layer with the softmax function:
		\begin{equation}
			\bm{P} = {\rm softmax} (\bm{W} \bm{R}^l + \bm{1}  \bm{b}^{\mathsf{T}}),
		\end{equation}
		where $\bm{W} \in \mathbb{R}^{4d \times c} , \bm{b} \in \mathbb{R}^{c} $ denotes learning parameters, $\bm{1} \in \mathbb{R}^{c} $ denotes the vector with all 1s  
		and $c$ denotes the number of the tags. 
		
		For a sequence of prediction $\bm{y} = (y_1, \cdots, y_m)$, 
		its  score is defined by:
		\begin{equation}
			score(\bm{s},\bm{y})=\sum_{0}^{m} T_{y_i, y_{i+1}} + \sum_{1}^{m} P_{ij},
		\end{equation}
		where 
		$T_{y_i, y_{i+1}}$ denotes the score of a transition from tag $y_i$ to tag $y_{i+1}$, 
		and $P_{ij}$ denotes the status score that the $i$th word is tagged $y_i$.


		\section{Experiments}
		\label{sec:experiments}
		\subsection{Experiments settings}
		\label{settings}
		
		\begin{tabhere}
			\caption{The statistics of datasets}
			\hspace*{-2.2em}
			\setlength{\tabcolsep}{2.5mm}{
			\label{tab:dataset}
			\begin{tabular}{llrrr}
				\toprule
				Language &  Dataset &  Train &  Dev &  Test \\ 
				\midrule
				German& CoNLL-2003& 12,705 & 3,068 & 3,160\\
				Spanish& CoNLL-2002&8,323&1,915&1,517\\
				Chinese& OntoNotes 4.0&15,509&4,405&4,462\\
				Chinese& Weibo & 1,350 & 270 & 270\\
				\bottomrule
			\end{tabular}
			}
		\end{tabhere}
		
		\textbf{Datasets.} 
		Experiments on four standard datasets are carried out to evaluate the proposed algorithm on the NER task. 
		These  standard datasets include CoNLL 2003 German~\cite{sang2003introduction}, 
		CoNLL 2002 Spanish~\cite{tksintro2002conll}, OntoNotes 4~\cite{weischedel2011ontonotes},  
		and Weibo NER~\cite{peng2015named}, where the last two are Chinese datasets. 
		The datasets from different language families are involved to prove the effectiveness of the proposed method. 
		We follow~\cite{che2013named} to select a part of OntoNotes 4 as a NER dataset. 
		Table~\ref{tab:dataset} shows the detailed statistics of these datasets.
		All the annotations are mapped to the BIOES format. 
		These datasets are chosen from different domains. 
		The CoNLL 2003 German, CoNLL 2002 Spanish,  and OntoNotes 4 datasets are in the news domain 
		while the WeiBo dataset is drawn from the social media website. 
		
		
		\textbf{Experimental Setup.} 
		We implement the base BiLSTM-CRF model using the PyTorch framework 
		and follow the configurations in~\cite{huang2015bidirectional} for comparative evaluation. 
		FastText embeddings\footnotemark[4]
		\footnotetext[4]{https://github.com/facebookresearch/fastText} are used to serve as basic word embeddings. 
		The pretrained static word embeddings, such as FastText, lose syntactic information in the sentences. 
		Therefore, the pretrained contextual word embeddings~\cite{devlin2019bert} are also utilized. 
		The translation module is implemented by Fairseq\footnotemark[5].
		\footnotetext[5]{https://github.com/pytorch/fairseq}
		The translation modules are trained on German-English, Spanish-English, and Chinese-English corpora in United Nations Parallel Corpus, respectively. 
		The pretrained English NER module employs the default NER model of Flair\footnotemark[6]
		\footnotetext[6]{https://github.com/zalandoresearch/flair}.
		
		We train the proposed model using stochastic gradient descent with no momentum for 
		150 epochs\footnotemark[7],
		\footnotetext[7]{These models are trained on two Nvidia GTX 1080Ti graphic cards.}
		with an initial learning rate of 0.1 and a learning rate annealing method 
		in which the training loss does not fall in 3 consecutive epochs. 
		The hidden size of BiLSTM is set to 256 and the layer of BiLSTM is set to 1. 
		With different datasets, we choose the learning rate  from $ \{0.025,0.05,0.1\}$ 
		and the batch size from $ \{8,16,32\}$. 
		All parameters are chosen by performance on the validation set. 
		Dropout is applied to word embeddings with a rate of 0.1 and to BiLSTM with a rate of 0.05, 
		which follows the recommendations in~\cite{reimers2017reporting}. 
		The hidden state size of the pretrained English NER model is also 256. 
		The aligned high-resource semantic feature has the same size as the hidden state of the pretrained English NER model. 
		The attention matrices in the first and last attention layers are recorded as transformation matrices, respectively.
		All experiments are repeated 5 times with different random seeds, 
		and we report average performance on the test set as the final performance.

		\subsection{Experiments on Indo-European languages}
		\label{Indo-European}
		
		In this subsection, we choose two of the Indo-European languages, German and Spanish, to evaluate the performance of the proposed method. Both low-resource languages, i.e, German and Spanish, belong to the same language family as English, which is the largest language family in the world. All labels are modified from BIO format to BIOES format following~\cite{akbik2018contextual}. 
		Since the translation model has multiple encoder-decoder attention layers, 
		two representative layers (the first layer and the last layer) are chosen in the proposed method 
		and the impact of different layers on CAN is explored in Subsection~\ref{sub:attenion-layer}. 
		$\rm CAN_{first}$ and $\rm CAN_{last}$ denotes that the proposed model exploits the attention matrices in the first and last attention layers to transfer the high-resource semantic features, respectively.

		\begin{tabhere}
			\caption{Results on German and Spanish NER datasets}
			\label{tab:german-spanish}
			\hspace*{-2em}
			\setlength{\tabcolsep}{3.8mm}{
			\begin{tabular}{lcc}
				\toprule
				Methods &  German &  Spanish \\
				\midrule
				\cite{gillick2016multilingual}  &-&82.95\\
				\cite{lample2016neural} 		&78.76&85.75  \\
				\cite{akbik2018contextual}	    &88.32 &-   \\
				\midrule
				BiLSTM+CRF						& 81.41&82.49\\
				\quad+$\rm CAN_{first}$				&82.23&82.73\\
				\quad+$\rm CAN_{last}$					& 82.45&84.23\\
				\midrule
				CharLM+BiLSTM+CRF				& 88.21&87.33\\	
				\quad+$\rm CAN_{first}$				&88.20&87.43  \\
				\quad+$\rm CAN_{last}$				&\textbf{88.41}&\textbf{88.16}\\
				\bottomrule
			\end{tabular}}
		\end{tabhere}

		Experimental results on German and Spanish datasets are shown in Table~\ref{tab:german-spanish}. 
		The evaluation metric is F1-score. 
		{BiLSTM+CRF} denotes the BiLSTM-CRF model with FastText embedding, which is treated as a baseline model. 
		The baseline model achieves a F1-score of 81.41\% and 82.49\% on German and Spanish datasets, respectively.
		The F1-scores of the models equipped with $\rm CAN_{first}$ ($\rm CAN_{last}$) are increased 
		to 82.23\% (82.65\%) and 82.73\% (84.23\%) on German and Spanish datasets, respectively.
		These performance improvements indicate the effectiveness of CAN.
		
		Existing state-of-the-art methods utilize language models to produce contextual embeddings, 
		which are orthogonal to our work. 
		{CharLM+BiLSTM+CRF} denotes the BiLSTM-CRF model with the character-level language model to generate contextual word embedding, 
		which is regarded as another baseline model.
		It achieves a F1-score of 88.21\% and 87.33\% on German and Spanish datasets, respectively.
		The performance of {CharLM+BiLSTM+CRF} overtakes consistently {BiLSTM+CRF}, 
		indicating the effectiveness of the character-level language model.
		{CharLM+BiLSTM+CRF+$\rm CAN_{first}$} achieves comparative F1-scores, i.e., 88.20\% and 87.43\%, 
		and {CharLM+BiLSTM+CRF+$\rm CAN_{last}$} obtains better F-scores, i.e., 88.41\% and 88.16\% on German and Spanish datasets, respectively. 
		These performance improvements indicate once again the effectiveness of CAN.

		Based on {BiLSTM+CRF} and {CharLM+BiLSTM+CRF}, CAN  further improves the performance 
		in seven out of eight cases except for {CharLM+BiLSTM+CRF+$\rm CAN_{first}$} on German dataset. 
		We attribute the effectiveness of CAN  to the high-resource semantic features that CAN obtains.
		Besides, we observe that CAN with the last attention matrix performs better than with the first attention matrix.
		This may be due to the fact that an attention matrix in a higher layer can capture more semantic dependency~\cite{chaudhari2019attentive}.

		
		\subsection{Experiments on Sino-Tibetan languages}
		\label{Sino-Tibetan}

		\begin{tabhere}
			\centering
			\caption{ Evaluation on OntoNotes 4.0. Gold Seg and No Seg denote whether or not to use the word segmentation, respectively}
			\label{tab:ontonotes}
			\hspace*{-0.5em}
			\setlength{\tabcolsep}{1.5mm}{
			\begin{tabular}{llccc}
				\toprule
				Input&  Methods & P(\%) &  R(\%) &  F1(\%) \\ 
				\multirow{4}{*}{Gold Seg}
				&\cite{yang2016combining} & 65.59&71.84& 68.57\\
				&\cite{yang2016combining} & 72.98&80.15& 76.40\\
				&\cite{che2013named} 	  &77.71&72.51& 75.02\\
				&\cite{wang2013effective} &76.43&72.32&74.32\\
				\midrule
				
				\multirow{6}{*}{No Seg}
				
				&\cite{zhang2018chinese} &76.35&71.56&  73.88\\
				&Char+BiLSTM+CRF&69.51&53.17& 60.25\\
				&\quad+$\rm CAN_{first}$&74.54&61.09& 67.15\\
				&\quad+$\rm CAN_{last}$&75.74&68.59& 71.99 \\
				&\quad+BERT+$\rm CAN_{first}$&78.12&78.36& 78.24 \\
				&\quad+BERT+$\rm CAN_{last}$&\textbf{80.42}&\textbf{82.02}& \textbf{81.21} \\
				\bottomrule
			\end{tabular}}
		\end{tabhere}
		
		In the previous subsection, the effectiveness of the proposed method is confirmed 
		on German and Spanish  which belong to the same language family as English. 
		In this subsection, to explore the generalization of the proposed method on other language families, 
		we focus on the second largest language family, i.e., the Sino-Tibetan language family. 
		This language family is distinct from the Indo-European language family, 
		and Chinese is the most widely used Sino-Tibetan language. 
		Unlike English, Chinese sentences comprise characters, and no space split these characters. 
		Word-based models need to first split sentences into words if there is no word explicitly segmented in sentences, 
		which will bring some inevitable errors. 
		Therefore, only the character-level embedding based methods are considered in this subsection 
		when no word is explicitly segmented in sentences. 
		
		\textbf{OntoNotes 4.0.} Table~\ref{tab:ontonotes} presents the results on Chinese OntoNotes 4.0. 
		Evaluation metrics are precision, recall, and F1-score. 
		For the OntoNotes dataset, gold-standard segmentation is available. 
		Existing state-of-the-art results are achieved by~\cite{yang2016combining}, 
		with gold-standard segmentation, discrete features, and semi-supervised data. 
		Using the baseline model (i.e., {Char+BiLSTM+CRF}), 
		the performance on Chinese OntoNotes 4.0 without segmentation is relatively lower. 
		The F1-score of {Char+BiLSTM+CRF} is 60.25\%. 
		Equipping $\rm CAN_{first}$ ($\rm CAN_{last}$) to the baseline model results in 
		an increase from 60.25\% to 67.15\% (71.99\%), 
		which indicates the effectiveness of the proposed CAN is once again in a distinct language family.
		Similarly, {Char+BiLSTM+CRF+$\rm CAN_{last}$} outperforms {Char+BiLSTM+CRF+$\rm CAN_{first}$}, 
		which is consistent with the aforementioned assumption that an attention matrix in a higher layer can capture more semantic dependency~\cite{chaudhari2019attentive}.
		In order to further improve the performance, we use the BERT model~\cite{devlin2019bert} to produce character embedding. 
		Our best model {Char+BiLSTM+CRF+BERT+$\rm CAN_{last}$} yields 81.21\% F1-score with no segmentation, 
		which outperforms previous state-of-the-art methods with no segmentation. 
		
		\end{multicols}
		
		\begin{tabhere}
			\centering
			\caption{Results on Weibo NER. NE, NM, and Overall denote named entities, nominal entities, and both of them, respectively}
			\label{tab:weibo}
			\setlength{\tabcolsep}{2.9mm}{
			\begin{tabular}{l|ccc|ccc|ccc}
				\toprule
				\multirow{2}{*}{Methods} & \multicolumn{3}{c|}{NE} & \multicolumn{3}{c|}{ NM}& \multicolumn{3}{c}{ Overall}\\
				&\bf P(\%)  & \bf R(\%)  & \bf F1(\%)  &\bf P(\%)  & \bf R(\%)  & \bf F1(\%)  &\bf P(\%) & \bf R(\%)  & \bf F1(\%) \\ 
				\midrule		
				Peng and Dredze\cite{peng2016improving}&66.67&47.22&55.28&74.48&54.55& 62.97&-&-&  58.99\\
				He and Sun\cite{he2017unified} &61.68&48.82&54.50&74.13&53.54&62.17&-&-&58.23\\
				Zhang and Yang\cite{zhang2018chinese} &-&-& 53.04&- &-&62.25&-&-&  58.79\\
				\midrule
				Char+BiLSTM+CRF&60.55&22.07&32.35&58.70&21.91& 31.91&59.55&22.05&32.18\\
				\quad+$\rm CAN_{first}$&61.15&23.67&34.13&58.62&25.12&  35.17&60.73&24.25&34.66\\
				\quad+$\rm CAN_{last}$&60.72&31.57&41.54&58.81&30.62&  40.27&61.21&30.96&41.12\\
				\midrule
				BERT+BiLSTM+CRF&72.97&67.71& 70.24&76.96&61.73&68.51&\textbf{74.52}&65.17&69.53\\			
				\quad+$\rm CAN_{first}$&72.99 &68.85&70.86&\textbf{77.19}&62.66& 69.17&{74.30}&66.37& 70.11 \\
				\quad+$\rm CAN_{last}$& \textbf{73.50}&\textbf{71.45}&\textbf{72.46}&77.01&\textbf{64.99}& \textbf{70.49}&74.17&\textbf{69.91}& \textbf{71.98} \\
				\bottomrule
			\end{tabular}}
		\end{tabhere}
		
		\begin{multicols}{2}
		\textbf{Weibo.} Results on the Weibo dataset are shown in Table~\ref{tab:weibo}, 
		where NE, NM, and Overall denote named entities, nominal entities, and both of them, respectively. 
		Evaluation metrics are precision, recall, and F1-score. 
		The previous model~\cite{peng2016improving} explored cross-domain data for semi-supervised learning. 
		As the dataset is too small, the baseline model (i.e., {Char+BiLSTM+CRF}) gives 32.35\%, 31.91\%, and 32.18\% F1-scores 
		on NE, NM, and Overall without external data and manual features, respectively. 
		Using the aligned high-resource semantic features captured by CAN, 
		{Char+BiLSTM+CRF+$\rm CAN_{last}$} achieves consistent and significant improvements on NE (41.51\%), NM (40.27\%), and Overall (41.12\%). 
		The performance comparison between the two versions  ($\rm CAN_{first}$ and $\rm CAN_{last}$) of CAN is consistent with that on OntoNotes. 
		{Char+BiLSTM+CRF+$\rm CAN_{last}$} yields 8.94\% improvement on average, in terms of F1-score, compared to {Char+BiLSTM+CRF}. 
		
		Moreover, we also explore the effectiveness of CAN with  contextual embeddings (i.e., BERT), which have rich syntactic information. 
		Utilizing BERT to produce character embedding, 
		the model {BERT+BiLSTM+CRF}, which serves as a baseline model, 
		achieves 70.24\%, 68.51\%, and 69.14\% F1-scores on NE, NM, and Overall, respectively. 
		 {BERT+BiLSTM+CRF+$\rm CAN_{last}$} achieves the F1-scores of 72.16\%, 70.09\%, and 71.28\% on NE, NM, and Overall, respectively. 
		It can be seen that the performance gets consistent improvements with the aligned high-resource semantic features obtained by CAN,
		which indicates the effectiveness of CAN.


	\section{Detailed Analyses}
	\label{sec:analysis}
	\subsection{CAN embedding}
	\label{embeddings}
	
	%
	To assess the aligned high-resource semantic features obtained by CAN, 
	we regard the aligned high-resource semantic features as special embeddings, named CAN embedding. 
	The performance of BiLSTM-CRF on a NER task is compared among three different embeddings: 
	random embeddings, FastText embeddings, and the proposed CAN embeddings.
	The random embeddings are to randomly generate a vector for each character 
	and the same character has the same embedding. 
	The random embeddings do not contain any syntactic or semantic information. 
	The FastText embeddings~\cite{bojanowski2017enriching} use the morphology and $n$-grams of words to represent words, 
	which is an efficient morphological representation. 
	The CAN embeddings denote the aligned high-resource semantic features 
	obtained by CAN. 
	Similarly, CAN embeddings with the attention matrices of the first and last attention layers are denoted as 
	$\rm CAN_{first}$ embeddings and  $\rm CAN_{last}$ embeddings, respectively.
	
	\begin{tabhere}
		\centering
		\caption{Comparison of different embeddings on the Weibo dataset using the same model BiLSTM-CRF}
		\setlength{\tabcolsep}{5mm}{
		\label{tab:embedding}
		\begin{tabular}{lccc}
			\toprule
			Embeddings & P(\%)  &  R(\%)  &  F1(\%)  \\ 
			\midrule
			Random  &36.12&10.55& 16.33\\  
			FastText &\textbf{59.55}&22.05& 32.18 \\	
			$\rm CAN_{first}$ &50.01&23.66& 32.12 \\ 
			$\rm CAN_{last}$ &57.22&\textbf{26.09}& \textbf{35.84} \\ 
			\bottomrule
		\end{tabular}}
	\end{tabhere}
	
	Experimental results are shown in Table~\ref{tab:embedding}. 
	The FastText embeddings and both CAN ($\rm CAN_{first}$ and $\rm CAN_{last}$) embeddings significantly outperform the random embeddings. 
	The $\rm CAN_{first}$ embeddings (32.12\%) perform comparably with the FastText embeddings (32.18\%). 
	The $\rm CAN_{last}$ embeddings obtain 35.84\% F1-score, which is 3.66\% higher than the FastText embeddings. 
	These experimental results indicate the effectiveness of the CAN embeddings, 
	which transfer high-resource language knowledge using aligned high-resource semantic features. 
	Furthermore, the CAN embeddings, especially $\rm CAN_{last}$, achieve better recall than the FastText embeddings, 
	which may be  due to the fact that CAN embeddings capture the task-specific information implied 
	in the pretrained high-resource NER model 
	and thus help {BiLSTM+CRF} find more named entities in the low-resource language.
	Meanwhile, it is also consistent with the previous work~\cite{liu2019linguistic} 
	that indicates the representations of higher layers of NLP models are more task-specific.  
	These experimental results illustrate that 
	the aligned high-resource semantic features obtained by CAN  transfer task-specific information from the pretrained high-resource NER model 
	to the low-resource one. 
	
	\subsection{Impact of different attention layers}
	\label{sub:attenion-layer}
	
	\begin{figurehere}
		\hspace*{-2.5em}
		\includegraphics[width=.5\textwidth]{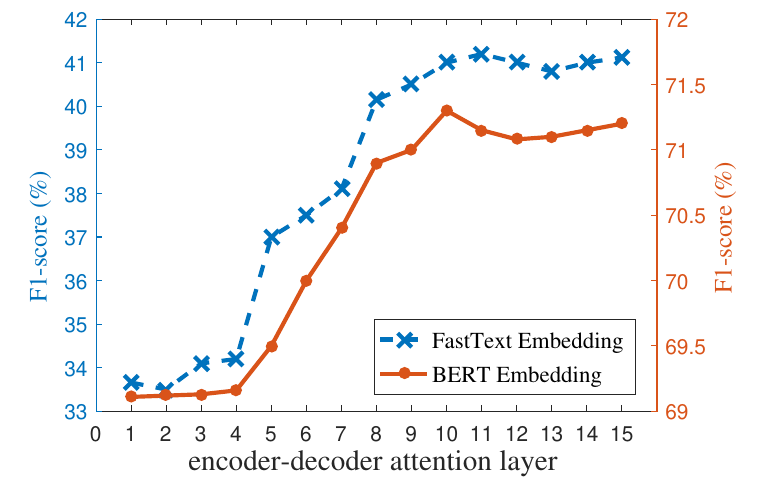}
		\caption{The performance of BiLSTM-CRF on the Weibo dataset with high-resource semantic features transferred from attention matrices in different attention layers.
			The dotted line denotes the performance of CAN with FastText embedding. 
			The solid line denotes the performance of CAN with BERT embedding.} 
		\label{fig:attention-layers}
	\end{figurehere}

	In this subsection, the aligned high-resource semantic features transferred 
	from attention matrices in different attention layers are studied to better understand and utilize CAN.
	An assumption is that attention matrices of higher layers in the neural machine translation model 
	capture deeper syntactic and semantic information. 
	The attention matrices of fifteen attention layers in the neural machine translation model 
	are recorded in the process of translating the Weibo dataset into English. 
	
	
	Figure~\ref{fig:attention-layers} presents the performance  of BiLSTM-CRF 
	using  aligned high-resource semantic features transferred from  attention matrices in different attention layers.
	As the layer of attention matrix used in CAN increases, 
	the performance shows an upward trend.
	When the attention matrices of the first to fourth attention layers are utilized to transfer the high-resource semantic features, 
	the performance has small improvements. 
	One potential reason is that the attention matrices in rather lower layers capture shallower syntactic and semantic information.
	When the attention matrices of the eighth or higher attention layers are used, the performance is good and smooth. 
	The optimal performance is achieved by using the tenth or eleventh attention layer. 
	This indicates that there is no need to rely too much on the depth of attention layers used in CAN.

	\section{Conclusion and Future Work}
	\label{sec:conclusion}
	In this paper, we seek to improve the performance of low-resource NER 
	by leveraging a pretrained high-resource English NER model. 
	This is achieved by the proposed method, converse attention network (CAN). 
	CAN  aligns and transfers semantic features in a pretrained model of the high-resource language to  low-resource languages.
	Extensive  experiments empirically indicate that, 
	CAN achieves consistent and significant performance improvements on multiple baseline models and datasets.
	This is of great practical importance for low-resource language datasets. 
	For future work, we would like to extend the proposed method to other NLP tasks, e.g., relation extraction and coreference resolution.
	
%
%

	\fontsize{8pt}{10.2pt}\selectfont
	\bibliographystyle{unsrt}
	\bibliography{Reference.bib}
	
	\end{multicols}
\newpage
~
\end{document}